\documentclass[10pt, journal, twoside]{ieeetran}
\usepackage{amsmath,epsfig}
\usepackage{setspace}
\usepackage{bm}
\usepackage{cite}
\usepackage{color}
\usepackage{float} 
\usepackage{array}
\usepackage{bbding}
\usepackage{pifont}
\usepackage{amssymb}
\usepackage{wasysym}
\usepackage{pdfpages}
\usepackage{epstopdf}
\usepackage{verbatim}
\usepackage{threeparttable}
\usepackage[colorlinks, linkcolor=red, anchorcolor=blue, citecolor=green]{hyperref}
\usepackage{lipsum,cite,multicol,multirow,amssymb,graphicx,array,epstopdf,subfigure,makecell,color, algorithm, algorithmic,booktabs}

\ifCLASSINFOpdf
\else
\usepackage[dvips]{graphicx}
\fi
\usepackage{url}
\usepackage{graphicx}

\begin{document}
\title{Single Underwater Image Enhancement Using\\ an Analysis-Synthesis Network}
	\author{Zhengyong~Wang, Liquan~Shen, Mei~Yu, Yufei~Lin and Qiuyu~Zhu}
\markboth{Journal of \LaTeX\ Class Files,~Vol.~14, No.~8, August~2015}%
{Shell \MakeLowercase{\textit{et al.}}: Bare Demo of IEEEtran.cls for IEEE Journals}
\maketitle

\begin{abstract} 
Most deep models for underwater image enhancement resort to training on synthetic datasets based on underwater image formation models. Although promising performances have been achieved, they are still limited by two problems: (1) existing underwater image synthesis models have an intrinsic limitation, in which the homogeneous ambient light is usually randomly generated and many important dependencies are ignored, and thus the synthesized training data cannot adequately express characteristics of real underwater environments; (2) most of deep models disregard lots of favorable underwater priors and heavily rely on training data, which extensively limits their application ranges. To address these limitations, a new underwater synthetic dataset is first established, in which a revised ambient light synthesis equation is embedded. The revised equation explicitly defines the complex mathematical relationship among intensity values of the ambient light in RGB channels and many dependencies such as surface-object depth, water types, etc, which helps to better simulate real underwater scene appearances. Secondly, a unified framework is proposed, named ANA-SYN, which can effectively enhance underwater images under collaborations of priors (underwater domain knowledge) and data information (underwater distortion distribution). The proposed framework includes an analysis network and a synthesis network, one for priors exploration and another for priors integration. To exploit more accurate priors, the significance of each prior for the input image is explored in the analysis network and an adaptive weighting module is designed to dynamically recalibrate them. Meanwhile, a novel prior guidance module is introduced in the synthesis network, which effectively aggregates the prior and data features and thus provides better hybrid information to perform the more reasonable image enhancement. Extensive experiments on synthetic data and five real underwater benchmarks demonstrate that our method achieves superior performance against other state-of-the-arts in underwater image enhancement. The synthetic dataset and code are publicly available at: https://github.com/Underwater-Lab-SHU/ANA-SYN.
\end{abstract}

\section{Introduction}
\begin{figure}[!t]
	\centering
	\centerline{\includegraphics[width=8.8cm, height= 4.5cm]{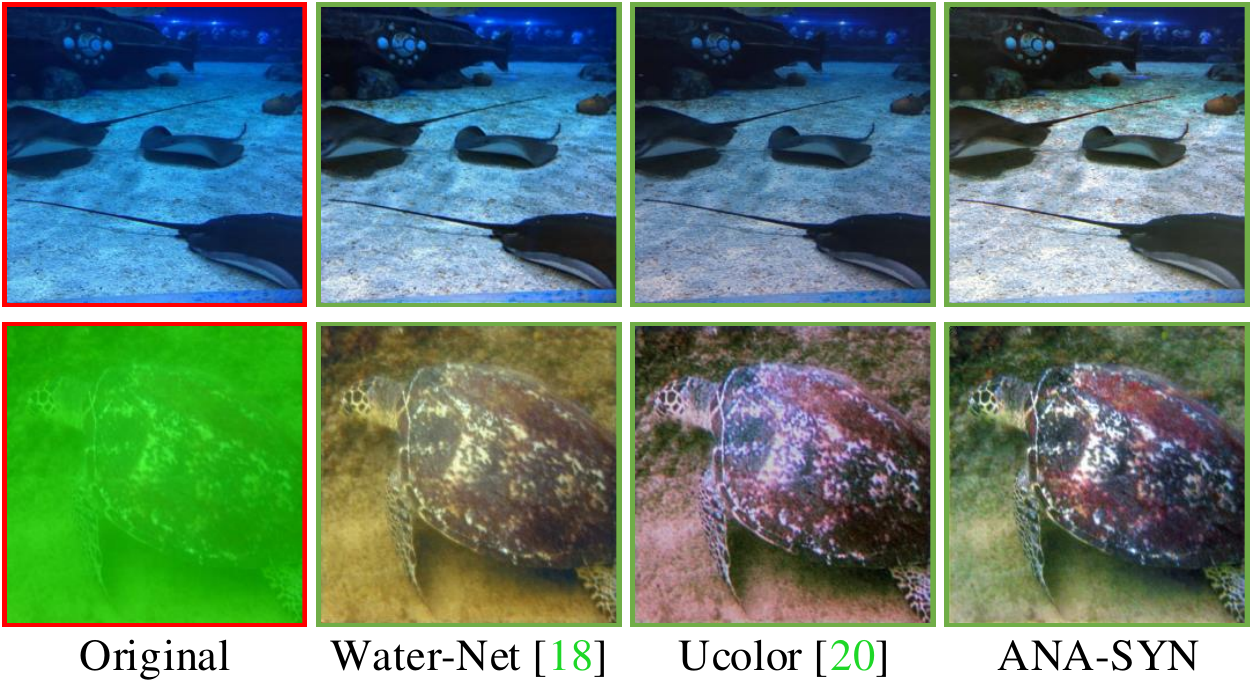}}
	\caption{Visual comparisons on two real underwater images. Our ANA-SYN not only effectively corrects color casts but also removes the haze on images, which is credited to the combination of priors and data information. In contrast, the competing methods cannot achieve satisfactory results. They introduce color artifacts and still remain haze in their enhanced results.}
	\label{our_result}
\end{figure}

\IEEEPARstart{U}{nderwater} image enhancement (UIE) aims to generate the clean version from a distorted underwater image, which is essential for many underwater applications such as biology, archaeology, underwater robotics and infrastructure inspection, etc. As shown in Fig.\ref{our_result}, the major problems in underwater images are low contrast, color casts and blurry details due to wavelength-dependent light absorption and scattering. When travelling through water, the lights with different wavelengths have different levels of attenuation ratio, leading to various degrees of color casts. Underwater images often appear to have bluish or greenish color tone since the red light having longer wavelength is absorbed more than the green and blue one. In addition, small particles in the water absorb most of light energy and change the direction of light before the light reaches the camera, resulting in low contrast and haze-like effects. In the past few years, many excellent algorithms have been proposed to address these problems.

Generally, UIE methods can be roughly divided into three categories: model-free, model-based and data-driven methods. Among them, model-free methods~\cite{iqbal2010enhancing,ghani2015underwater,ghani2015enhancement,Fu2014,Zhang2017,ancuti2012enhancing,ancuti2017color,Ancuti2020} mainly address the pixels to enhance underwater images. These methods can improve contrast, brightness and saturation. However, they ignore the physical degradation process of underwater images and thus the visual quality improvement of enhanced results is limited. Differently, model-based methods~\cite{drews2016underwater, wang2017single, peng2017underwater, li2016underwater, Akkaynak2017, Akkaynak2018, Akkaynak2019} usually estimate key parameters of physical models by various hand-crafted priors to restore clean images, which perform well in some cases. However, these methods tend to make inaccurate parameters estimation and produce unsatisfactory results in some complex scenarios since their priors often fail. Recently, data-driven methods~\cite{li2018emerging, guo2019underwater, li2019underwater, Xue2021, Li2021} have been employed to estimate physical parameters or directly predict clear images. Usually, these approaches rely on a large number of real underwater images and their clean counterparts for training. However, it is impractical to acquire large quantities of the corresponding ground-truths in the underwater scenarios. Consequently, most data-driven methods heavily depend on synthetic datasets based on underwater image formation models to train their models. Although these methods have shown remarkable improvements, they still face two major challenges as follows. 

Firstly, existing underwater image synthesis models have an inherent limitation in synthesizing training samples, which cannot adequately express characteristics of real underwater images. They usually randomly generate the homogeneous global ambient light in the data synthesis process of various water types. This operation introduces significant errors since the intensity values of the ambient light in RGB channels are not independent and influenced by many dependencies, such as water types and surface-object depth, etc. In this paper, we establish a new underwater synthetic dataset using a revised ambient light synthesis equation that takes these dependencies into account. The revised equation clearly defines the complex mathematical relationship among intensity values of the ambient light in RGB channels, absorption coefficient, scattering coefficient and surface-object depth, which is more accurate in simulating the color, contrast and blurriness appearance of real underwater scenes.

\begin{figure}[!t]
	\centering
	\centerline{\includegraphics[width=8.6cm,height=4.0cm]{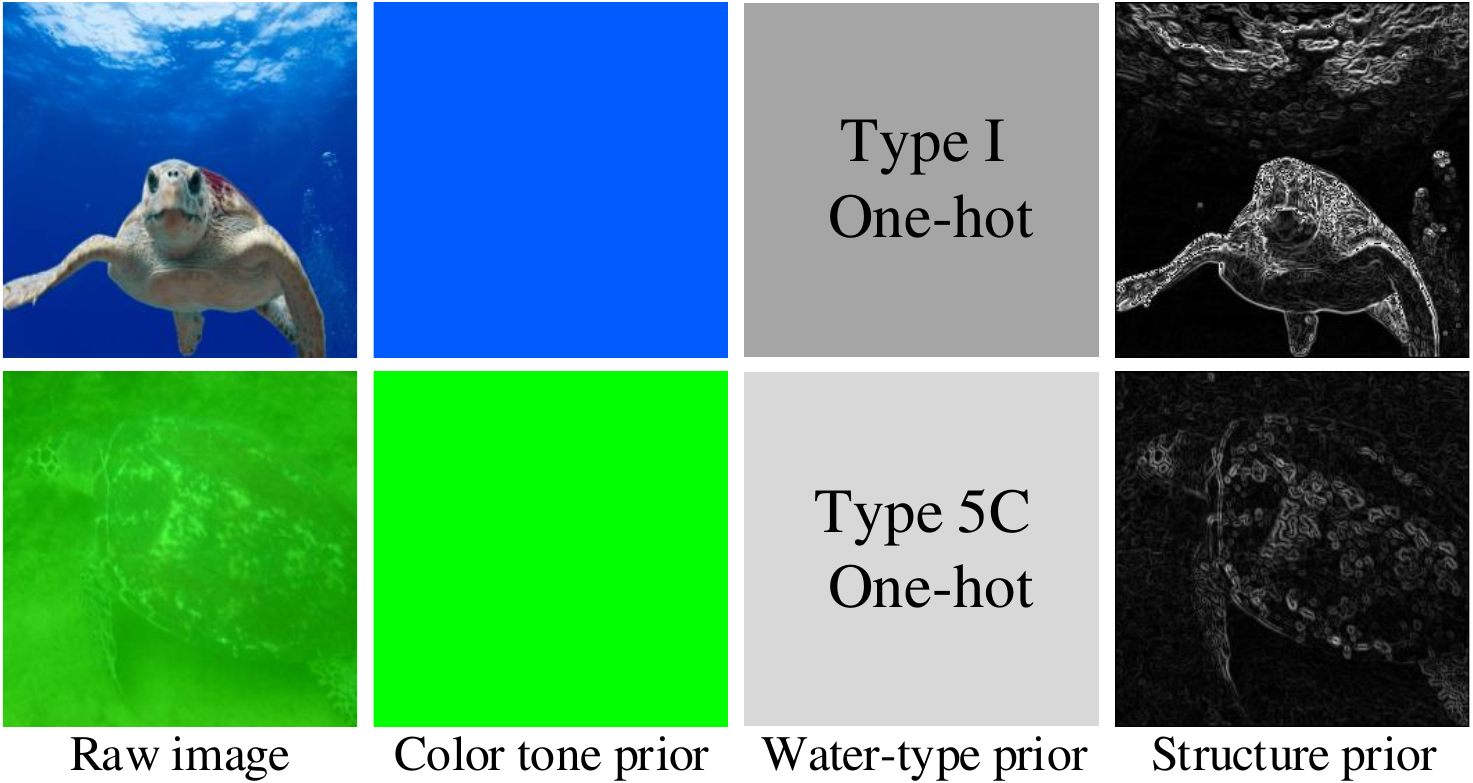}} 
	\caption{Underwater images and their corresponding three priors. It can be observed that these priors reflect some unique information contained in underwater images to some extent, such as color, texture and water-body.}
	\label{Prior_show}
\end{figure}

Secondly, most deep models designed with engineering experiences are highly dependent on the scale and quality of training data and thus their performances cannot be always guaranteed. There are lots of unique priors can be used in the underwater field to help networks learn more discriminative features, recover better images and improve the generalization of the model~\cite{liu2018learning}. Two examples are shown in Fig.\ref{Prior_show}, including the distorted images and three corresponding priors (i.e., color tone, water-type and structure prior). Color tones of underwater images can reflect the distortion information associated with color to some extent, which is beneficial for color correction. Similarly, the structural information can provide some high-frequency local features of edges, which guides networks to better restore texture and geometry. In addition, by incorporating the water-type prior embedded with water body information into the network, the learned model can achieve better robustness to different water body conditions.

\begin{figure}[!t]
	\centering
	\centerline{\includegraphics[width=8.6cm,height=4cm]{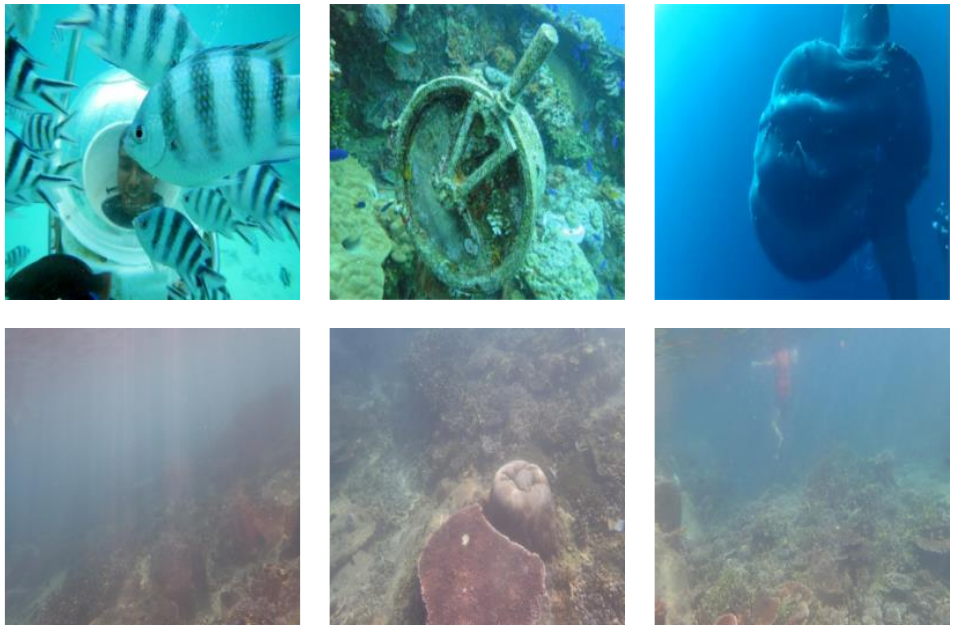}}
	\caption{Examples of real underwater images, which have obvious characteristics of underwater image quality degradation. It can be observed that the main distortion in the 1st row is mainly reflected in color casts, and the 2nd row is mainly reflected in blurry details.}
	\label{prior_3}
\end{figure}

Based on the above analysis, a novel analysis-synthesis framework is proposed for UIE tasks, called ANA-SYN, which effectively enhances underwater images under the combination of priors (underwater domain knowledge) and data information (underwater distortion distribution). The proposed framework consists of two parts, namely an analysis network and a synthesis network. Among them, analysis network is employed to explore some underwater image priors (e.g., color tone prior, water-type prior, structure prior, etc). However, for images with different degrees of distortion, the importance of each prior is not constant, which varies with underwater scenes. As shown in Fig.\ref{prior_3}, underwater images in the first row are mainly affected by color distortions, while images in the second row are with blur-dominated distortions. For the former, the color tone prior is more important and for the latter, the structure prior is more significant. Therefore, a novel adaptive weighting module is embedded to dynamically learn weight maps of each prior and recalibrate them into more accurate weighted priors. Synthesis network aggregates priors and data information to perform image enhancement, in which a new prior guidance module is introduced to adequately fuse prior and data features by a pair of learnable prior modulation parameters. As shown in Fig.\ref{our_result}, our ANA-SYN can effectively handle real underwater images with different types of quality degradation and produces much better results. 

The main contributions of this paper can be summarized as follows: 
\begin{enumerate}
	\item We build a new underwater synthetic dataset based on a revised ambient light synthesis equation. To the best of our knowledge, it is the first dataset that takes the relationship among the ambient light intensity values, water-type and surface-object depth into account during the data synthesis process, and the synthesized images are more realistic and natural.
	\item We design a new framework, named ANA-SYN, which effectively integrates underwater priors and data information for solving underwater image enhancement tasks, and performs well in real-world scenarios.
	\item We propose a novel adaptive weighting module in the analysis network, which can adaptively assign different weights to various priors based on the their importance to the input image, and recalibrate them into more accurate weighted priors.
	\item A prior guidance module is introduced in the synthesis network, which effectively fuses prior and data features by learning a pair of modulation parameters, guiding the meaningful interaction of priors and data information.
\end{enumerate}


\section{Related work}
\subsection{Underwater Image Synthetic Datasets}
One challenge faced with UIE tasks is lack of real datasets containing their corresponding ground-truths. Li~\textit{et al}.~\cite{li2019underwater} construct a real underwater image enhancement benchmark dataset, including 890 images and their respective references (not the actual ground truth). Liu~\textit{et al}.~\cite{liu2019real} build a real underwater dataset which consists of three subsets targeting at three challenging aspects, i.e, visibility degradation, color casts and higher-level tasks, respectively. Although these images are real and reliable, they usually have limited numbers, content and types of quality distortion. More importantly, they do not provide the corresponding ground-truth images. In general, it is impossible to obtain the corresponding ground-truth with respect to a real underwater image. Thus, most existing deep models resort to training on some synthetic datasets.

Underwater image synthesis is usually solved from two different perspectives: based on Generative Adversarial Networks (GAN)~\cite{li2017watergan, fabbri2018enhancing} and physical models~\cite{Li2020UnderwaterSP, ding2019jointly}. Li~\textit{et al}.~\cite{li2017watergan} propose a deep model, named WaterGAN, to generate underwater-like images from in-air images and depth maps in an unsupervised manner. Similarly, Fabbri~\textit{et al}.~\cite{fabbri2018enhancing} employ CycleGAN to generate distorted images from clean images based on weakly supervised distribution transfer. The GAN-based synthesis methods show a great advantage of generating underwater-like images in an unsupervised pipeline, enabling to transfer the distribution of images instead of all colors, contrast and blur details on the surface. However, these methods heavily depend on training samples, which are easy to produce unrealistic and unnatural artifacts.

Recently, several underwater image synthesis methods based on physical models are proposed. Li~\textit{et al}.~\cite{Li2020UnderwaterSP} establish an underwater synthetic image database via a simplified physical model, including ten subsets for different water types. Ding~\textit{et al}.~\cite{ding2019jointly} modify the work of~\cite{Li2020UnderwaterSP} by introducing the color shift and object blurriness effect, producing relatively good synthetic data. Nevertheless, the homogeneous ambient light is usually randomly generated in the process of synthesis, which makes a large difference between the synthetic and real underwater images. In this paper, the intensity value relationship of the ambient light in RGB channels at one depth point in the water is defined by a revised ambient light synthesis equation. By embedding the revised equation into the synthesis process, the synthesized data is closer to real underwater data.
\begin{figure*}[!t]
	\centering
	\centerline{\includegraphics[width=18.1cm,height=6.8cm]{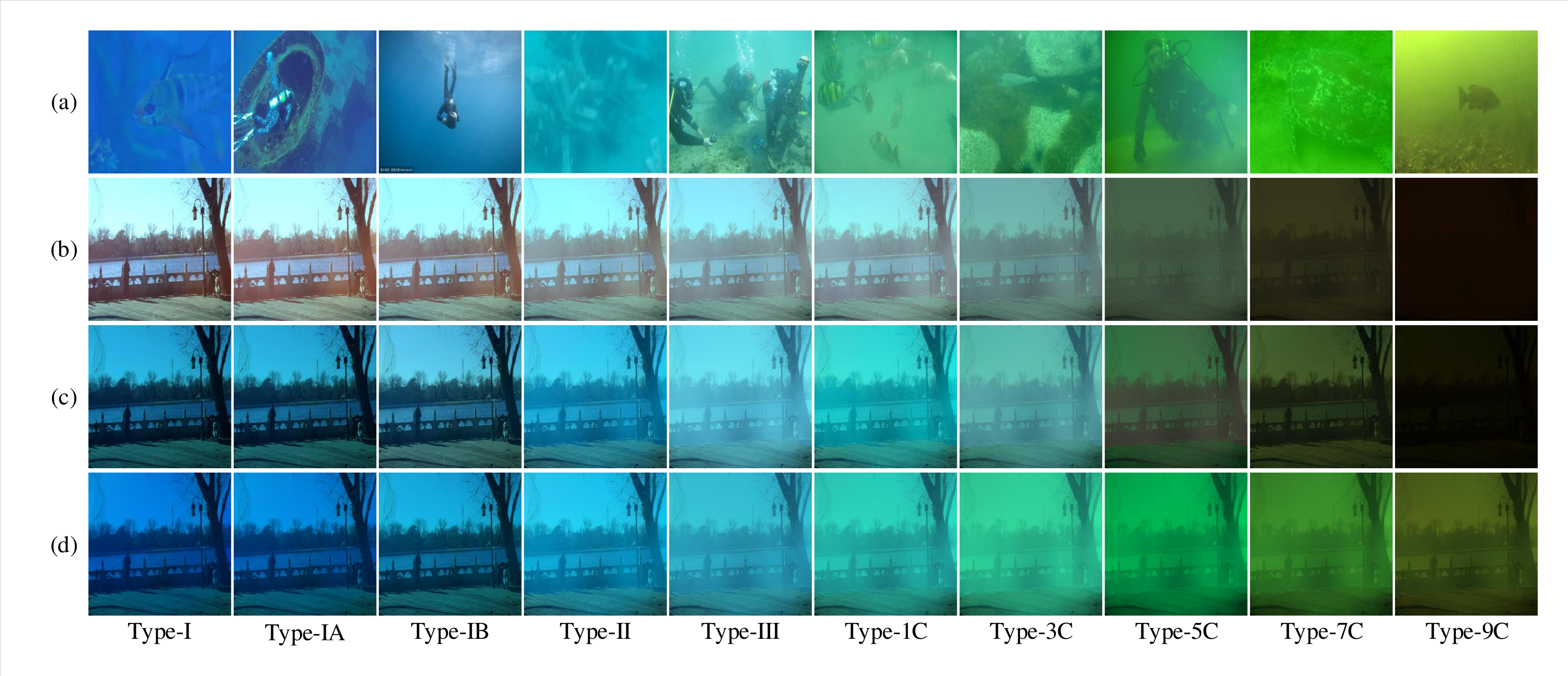}}  %
	\caption{Samples of the synthesized underwater images with different types. (a) real-world underwater images, (b) synthetic underwater-like images based on~\cite{Li2020UnderwaterSP}, (c) synthetic underwater-like images based on~\cite{ding2019jointly}, (d) synthetic underwater-like images using our proposed revised ambient light synthesis equation.}
	\label{Syn_dataset_3}
\end{figure*}

\subsection{Underwater Image Enhancement Methods}
Existing underwater image enhancement algorithms can be roughly classified into three types: model-free~\cite{iqbal2010enhancing,ghani2015underwater,ghani2015enhancement,Fu2014,Zhang2017,ancuti2012enhancing,ancuti2017color,Ancuti2020}, model-based~\cite{drews2016underwater, wang2017single, peng2017underwater, li2016underwater,Akkaynak2017,Akkaynak2018,Akkaynak2019} and data-driven methods~\cite{li2018emerging, guo2019underwater, li2019underwater, Xue2021, Li2021}.

Model-free methods mainly change image pixel values to improve visual quality of underwater images, such as pixels stretching and adjustment~\cite{iqbal2010enhancing,ghani2015underwater,ghani2015enhancement,Ancuti2020}, retinex decomposition~\cite{Fu2014,Zhang2017} and image fusion~\cite{ancuti2012enhancing, ancuti2017color}. For example, Ancuti~\textit{et al}.~\cite{ancuti2012enhancing} propose an underwater image enhancement method, in which a contrast-enhanced image and a color-corrected image are fused by a multi-scale mechanism to generate an enhanced image with better global contrast and detail information. Based on underwater optical imaging theory, Ancuti~\textit{et al}.~\cite{ancuti2017color} modify the color balance method to effectively fuse images, leading to more natural results. Recently, Ancuti~\textit{et al}.~\cite{Ancuti2020} design a color channel compensation pre-processing method based on the observation that the information contained in at least one color channel is close to completely lost under underwater, which can effectively remove color artifacts. Model-free methods can improve the contrast and saturation of underwater images to some extent. However, they only depend on human visual brightness and color perception, while ignoring the complexity of underwater scenarios. It is difficult for them to achieve promising results in images with complex underwater degraded environments and lighting conditions.

Model-based methods usually deduce some key parameters of the physical model via some hand-crafted priors and then recover clean images by inverting the degradation process. The priors include underwater dark channel prior~\cite{drews2016underwater}, attenuation curve prior~\cite{wang2017single}, blurriness prior~\cite{peng2017underwater} and minimum information prior~\cite{li2016underwater}, etc. For example, Peng~\textit{et al}.~\cite{peng2017underwater} propose a method to better estimate depth maps for underwater scenes based on the intrinsic characteristics of underwater image blurriness and light absorption. Li~\textit{et al}.~\cite{li2016underwater} integrate the minimum information loss and histogram distribution prior for depth estimation to effectively recover underwater images. Recently, Akkaynak~\textit{et al}. take many underwater dependencies into account and present a revised underwater image formation~\cite{Akkaynak2017,Akkaynak2018}, which is more physically accurate. From this model, they propose a color correction algorithm~\cite{Akkaynak2019} using underwater RGB-D images to restore images. Although these methods perform reality relatively in some cases, they heavily depend on hand-crafted priors. Thus, they tend to fail when hand-crafted priors are not valid on specific images.

With the advance of deep learning and large-scale synthetic datasets, data-driven methods have received significant attention in recent years. Current methods either design end-to-end modules, or utilize deep models directly to estimate physical parameters, and then restore the clean image based on the degradation model. To alleviate the need for paired training data, Li~\textit{et al}.~\cite{li2018emerging} develop an underwater weakly supervised learning method to enhance images. Guo~\textit{et al}.~\cite{guo2019underwater} design a multi-scale dense GAN which combines GAN loss with L1 and gradient loss to better learn the distribution of features for robust image enhancement. Recently, Li~\textit{et al}.~\cite{li2019underwater} employ three weights learned by a gate network to fuse three pre-processing versions of a distorted image for a better output. Furthermore, a new cross-color-spaces and cross-channel method is proposed in~\cite{Xue2021}, which realizes the disentanglement of color and haze effects to avoid unsatisfactory colors and blurring in some areas. 

Although data-driven methods have made significant progresses, current network architectures ignore lots of available underwater priors and are tightly dependent on training data. The generalization still falls behind some conventional methods and their performances are limited in the real world. Rich underwater priors can provide a proper guidance to the network and make the trained model more robust~\cite{liu2018learning}. In this paper, we effectively aggregate both underwater prior and data information to better solve UIE tasks. 

\section{Proposed Underwater Dataset}
In this section, we first analyze the limitations of existing underwater synthesis models, and propose a revised ambient light synthesis equation. Subsequently, the construction details of our new dataset are introduced.

\subsection{Limitations of existing underwater synthesis models}
In many studied models, the most widely one used to describe the degradation process of underwater images is Jaffe-McGlamery model~\cite{chiang2011underwater}. Mathematically, it is modeled as:
\begin{equation}\label{eq1}
\begin{aligned}
I_{c}(x)=J_{c}(x) \cdot e^{-\beta_{c}(\lambda) \cdot d(x)}+B_{c}^{\infty} \cdot \left(1-e^{-\beta_{c}(\lambda) \cdot d(x)}\right)
\end{aligned}
\end{equation}
where $x$ denotes a point in the underwater scene, $I_{c}(x)$ is the total signal received by the camera, ${J_{c}(x)}$ represents the clean image at point $x$, ${c=\{\mathrm{r}, \mathrm{g}, \mathrm{b}\}}$ denotes color channels, $d(x)$ is the object–camera distance, ${B_{c}^{\infty}}$ is the homogeneous ambient light and ${{\beta}_{c}(\lambda)}$ is the sum of absorption coefficient $a_{c}(\lambda)$ and scattering coefficient ${b_{c}(\lambda)}$~\cite{solonenko2015inherent}, i.e., ${{\beta}_{c}(\lambda)} = {a_{c}(\lambda)} + {b_{c}(\lambda)}$. Li~\textit{et al}.~\cite{Li2020UnderwaterSP} synthesize the first underwater-like dataset based on Eq.\ref{eq1}. Ding~\textit{et al}.~\cite{ding2019jointly} introduce the color component to improve the work of~\cite{Li2020UnderwaterSP}. $J_{c}(x)$ is redefined as the scene irradiance at point $x$, which is closer to the characteristic of light propagation in the water. 

When light propagates the surface-object distance $D(x)$ and reaches the underwater scene point $x$, the energy attenuation can be written as,
\begin{equation}\label{eq2}
J_{c}(x)=E_{c}^{i n}(x) \cdot e^{-a_{c}(\lambda) \cdot D(x)}
\end{equation}
where $E_{c}^{i n}(x)$ is the initial energy at point $x$ and $D(x)$ is the distance from the surface to the object, i.e. surface-object distance. Therefore, an underwater image formation model which takes account of color shift can be modeled as:
\begin{equation}\label{eq3}
\begin{aligned}
I_{c}(x)= E_{c}^{i n}(x)  &\cdot e^{-a_{c}(\lambda) \cdot D(x)} \cdot e^{-\beta_{c}(\lambda) \cdot d(x)} + B_{c}^{\infty} \\
& \cdot \left(1-e^{-\beta_{c}(\lambda) \cdot d(x)}\right), c \in\{\mathrm{r}, \mathrm{g}, \mathrm{b}\}
\end{aligned}
\end{equation}
\indent Ding~\textit{et al}.~\cite{ding2019jointly} establish a new underwater-like dataset based on Eq.\ref{eq3} by multiple parameters including absorption coefficient ${a_{c}(\lambda)}$, scattering coefficient ${b_{c}(\lambda)}$, object–camera distance $d(x)$, surface-object distance $D(x)$ and ambient light $B_{c}^{\infty}$. They deem that low visibility and color casts are key degradation parameters for image quality and ignore the effect of ambient light. Existing synthesis models usually randomly generate the homogeneous ambient light in the data synthesis process and ignore many important dependencies. 

To illustrate the drawback of existing synthesis models, ten different types of real underwater images and some synthetic images based on various synthesis models are shown Fig.\ref{Syn_dataset_3}. By comparing the 1st, 2nd and 3rd rows of Fig.\ref{Syn_dataset_3}, it can be clearly observed that the synthesized images based on~\cite{Li2020UnderwaterSP} and~\cite{ding2019jointly} cannot simulate the characteristics of real underwater images well, especially in color casts. In our opinion, this is mainly caused by incorrect ambient light settings. Although the ambient light values of RGB channels at point x in the underwater scene are constants, they are related to many dependencies including water-type, attenuation coefficients and surface-object distance~\cite{Akkaynak2017, Akkaynak2018, Akkaynak2019} since the light can be gradually absorbed with the distance from surface to object and the light absorption rate changes with the water types. Therefore, exploring more potential relationships among the ambient light intensity values in RGB channels and many dependencies is important to synthesize more accurate underwater-like data.

\subsection{Proposed Revised Ambient light Synthesis Equation}
In this section, a revised ambient light synthesis equation is developed. Assuming that there is a small water disk with thickness $dz$, radiation scattered by this small disk from all other directions can be written as~\cite{liu2019systematic}:
\begin{equation}\label{eq4}
dL(z,\lambda) = b(\lambda) \cdot E(D, \lambda)dz
\end{equation}
where $E(D,\lambda)$ is the ambient light value at the surface-object depth $D$, $L(z,\lambda)$ is the scene object and $b(\lambda)$ is the scattering coefficient. Based on Beer’s Law of exponential decay, the received radiance~\cite{schechner2004clear} at a distance $z$ is
\begin{equation}\label{eq5}
d B(z, \lambda)=d L(z, \lambda) \cdot e^{-\beta(\lambda) \cdot z}
\end{equation}

By substituting Eq.\ref{eq4} into Eq.\ref{eq5} and integrating $d$ from $z_{1}=0$ to $z_{2}=z$, the ambient light can be rewritten as
\begin{equation}
\label{eq6}
B(z, \lambda)=\frac{b(\lambda) \cdot E(D, \lambda)}{\beta(\lambda)}\left(1-e^{-\beta(\lambda) \cdot z}\right)
\end{equation}
when $z \rightarrow \infty$, the ambient light value at one point $x$ can be calculated~\cite{Akkaynak2017,Akkaynak2018,Akkaynak2019},
\begin{equation}
\label{eq7}
\begin{aligned}
B^{\infty}(\lambda)=&\frac{b(\lambda) \cdot E(D,\lambda)}{\beta(\lambda)}=\frac{b(\lambda) \cdot E(0, \lambda) \cdot e^{-\beta(\lambda) \cdot D}}{\beta(\lambda)}= \\
&\frac{b(\lambda) \cdot e^{-[a(\lambda)+b(\lambda)] \cdot D}}{a(\lambda)+b(\lambda)} \cdot E(0, \lambda)
\end{aligned}
\end{equation}
where $E(0,\lambda)$ is the ambient light at the sea surface. 

It can be observed that $B_{c}^{\infty}$ depends on the scattering and attenuation coefficients $a(\lambda)$ and $b(\lambda)$, range $D$ and ambient light value at the sea surface $E(0,\lambda)$. Thus, the intensity values of ambient light in RGB channels at one point in the underwater scene, ($B_{r}$, $B_{g}$, $B_{b}$) are,
\begin{equation}\label{eq8}
\begin{aligned}
B_{r} &=\frac{b(r) \cdot e^{-[a(r)+b(r)] \cdot D}}{a(r)+b(r)} \cdot E(0, r)
\end{aligned}
\end{equation}
\begin{equation}\label{eq9}
\begin{aligned}
B_{g} &=\frac{b(g) \cdot e^{-[a(g)+b(g)] \cdot D}}{a(g)+b(g)} \cdot E(0, g)
\end{aligned}
\end{equation}
\begin{equation}\label{eq10}
\begin{aligned}
B_{b} &=\frac{b(b) \cdot e^{-[a(b)+b(b)] \cdot D}}{a(b)+b(b)} \cdot E(0, b)
\end{aligned}
\end{equation}

The energy corresponding to red, green, blue channels above the water surface shall be the same, i.e., $E(0, r)=E(0, g)=E(0, b)$~\cite{chiang2011underwater}. The intensity ratios of ambient light values in RGB channels are,
\begin{equation}\label{eq11}
\begin{aligned}
\frac{B_{r}}{B_{g}}=\frac{b(r)}{b(g)} \cdot \frac{a(g)+b(g)}{a(r)+b(r)} \cdot e^{-[a(r)+b(r)-a(g)-b(g)] \cdot D}
\end{aligned}
\end{equation}
Similarly,
\begin{equation}\label{eq12}
\begin{aligned}
\frac{B_{b}}{B_{g}}=\frac{b(b)}{b(g)} \cdot \frac{a(g)+b(g)}{a(b)+b(b)} \cdot e^{-[a(b)+b(b)-a(g)-b(g)] \cdot D}
\end{aligned}
\end{equation}

Thus, the intensity values of the ambient light in RGB channels $B_{c}^{\infty}(\lambda)$ can be written as,
\begin{equation}\label{eq13}
\mathrm{B}_{c}^{\infty}(\lambda)=\left\{\frac{B_{r}}{B_{g}} \cdot B_{g}, B_{g}, \frac{B_{b}}{B_{g}} \cdot B_{g}\right\}
\end{equation}

In the data synthesis process, the ambient light value of one channel is set based on Eq.\ref{eq13}, and then the ambient light value of the other two channels can be directly calculated by the revised relationship. Compared with existing synthetic models that randomly generate the ambient light, our revised equation dynamically considers many dependencies and thus reduces distances between synthetic and real images, which makes the synthesized data better reflect the characteristics of real underwater images. During synthesis, the ambient light value in the green channel is empirically set as a random variable according to application needs, i.e., $0.5 \leq B_{g} \leq 1$~\cite{Li2020UnderwaterSP, ding2019jointly}.

\subsection{Dataset Construction Details}
A new underwater dataset is synthesized using NYU-depth2 RTTS dataset\footnote[1]{{https://sites.google.com/view/reside-dehaze-datasets/reside-v0}} based on Eq.\ref{eq3} and our proposed revised ambient light synthesis equation. The NYU-depth2 RTTS dataset includes 4322 clean indoor images and the corresponding scene depth maps. 1000 clean images are randomly chosen as ground-truths and the scene depths vary from 0.25m to 20m as object-camera distances, i.e. $0.25 \mathrm{m} \leq d(x) \leq 20 \mathrm{m}$. Besides, the surface-object distance is set from 0m to 5m, i.e. $0 \mathrm{m} \leq D(x) \leq 5 \mathrm{m}$. The absorption coefficient $a(\lambda)$ and scattering coefficient $b(\lambda)$ in the wavelength red (650nm), green (525nm) and blue (450nm) channels for one water type can be derived from~\cite{solonenko2015inherent}. 

Based on the description above, a new underwater synthetic dataset containing ten subsets (i.e., Type I, II, III, IA, IB, 1C, 3C, 5C, 7C and 9C) is built. Some synthetic samples with different synthesis models are shown in Fig.\ref{Syn_dataset_3}, where it is obvious that the appearance and features of our proposed dataset are closer to real underwater images compared to the synthesized datasets based on~\cite{Li2020UnderwaterSP} and~\cite{ding2019jointly}. In addition, we verify the effectiveness of the proposed dataset in Section~\ref{section:my} and more results are given in the supplementary material.
\label{section:me}

\begin{figure*}[htbp]
	\centering
	\centerline{\includegraphics[scale = 0.48]{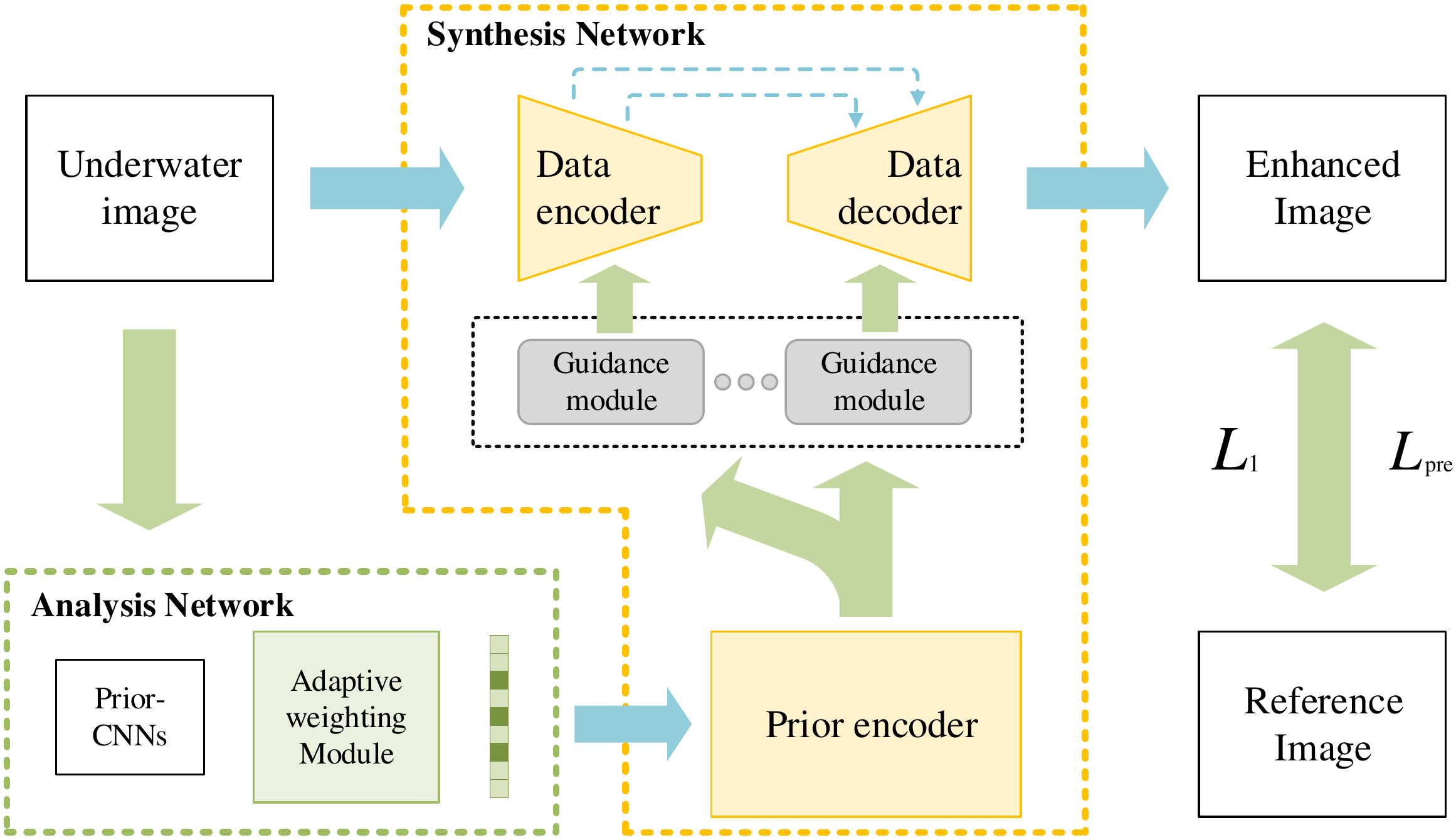}} 
	\caption{The structure of our ANA-SYN framework for underwater image enhancement. The framework includes an analysis network used for priors exploration and a synthesis network employed to priors integration. The details of analysis network and synthesis network are shown in Fig.\ref{Analysis Network} and Fig.\ref{Synthesis Network}, respectively.}
	\label{Method}
\end{figure*}

\begin{figure*}[!t]
	\centering
	\centerline{\includegraphics[scale = 0.76]{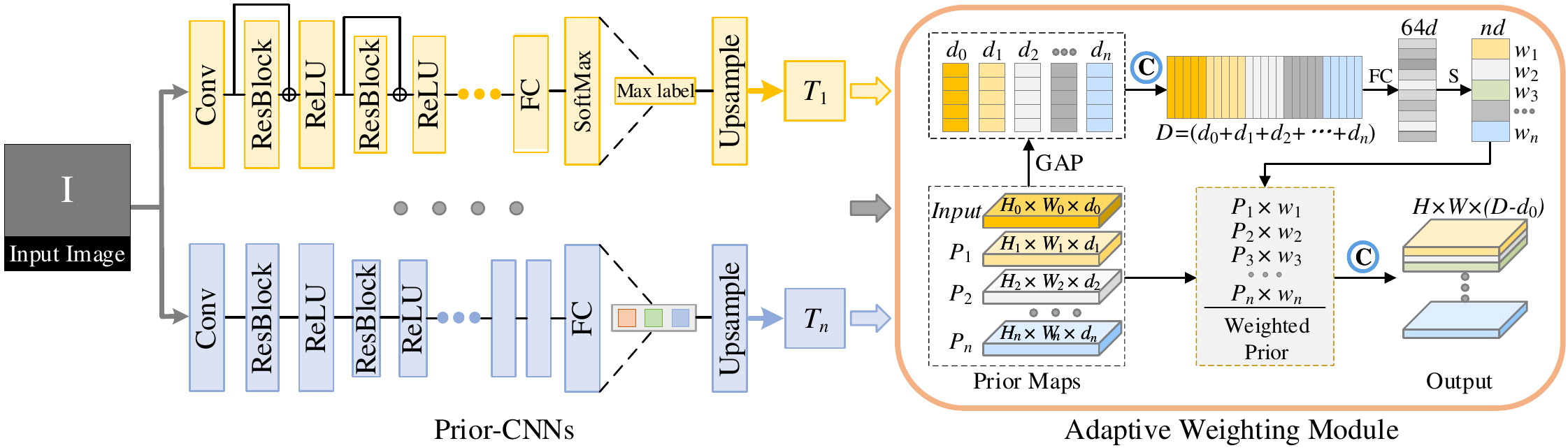}}
	\caption{Overview of our analysis network structure. Analysis network consists of two parts, a Prior-CNNs module and an adaptive weighting module. First, various priors of the input image are extracted by the frozen Prior-CNNs module, and then analysized by the adaptive weighting module, which predicts weight maps based on their importance of the input image. Finally, more accurate weighted priors can be obtained by adaptively recalibrating the prior using the corresponding weight map. In this paper, $n$ = 3.}
	\label{Analysis Network}
\end{figure*}

\section{Proposed ANA-SYN Framework}
In this paper, we aim to integrate rich underwater priors and data information to solve UIE tasks. A unified analysis-synthesis framework is designed, termed ANA-SYN, which contains two modules: an analysis network for priors exploitation and a synthesis network for priors integration. The overview of our ANA-SYN is shown in Fig.\ref{Method}. Specifically, analysis network first adopts a pre-trained prior  module (Prior-CNNs) to extract various underwater priors. Then, a novel adaptive weighting module is developed, which dynamically learns a set of prior-specific weights to recalibrate priors according to their usefulness for input image. Synthesis network takes the original image and weighted priors as inputs to perform image enhancement, in which a prior encoder and a data encoder-decoder are employed to extract prior features and data features, respectively. A new prior guidance module is exploited to effectively fuse prior and data features on each layer. More details are presented in the following subsections.

\subsection{Analysis Network}
Analysis network aims at obtaining and recalibrating some underwater priors, as shown in Fig.\ref{Analysis Network}. The network includes two parts: a pre-trained prior module (Prior-CNNs) for extracting various priors, followed by an adaptive weighting module for recalibrating them based on their importance for the input image. As described in Section \ref{section:me}, the synthesized dataset has some different underwater parameter pairs including color tones, water types, etc. Convolutional neural networks trained on these parameter pairs are employed to extract various priors from the input image, called Prior-CNNs. It is worth noting that these pre-trained networks can be designed based on some commonly used network structures (e.g., resnet-block and densenet-block, etc) and the parameters of pre-trained Prior-CNNs are kept frozen in the analysis network. More detailed architectures of priors extraction networks can be found in the supplementary material. In this paper, the color tone prior, water-type prior and structure prior are extracted to validate the applicability of our ANA-SYN framework (some examples of the extracted priors are shown in Fig.\ref{Prior_show}). 

\begin{figure*}[!t]
	\centering
	\centerline{\includegraphics[scale=0.70]{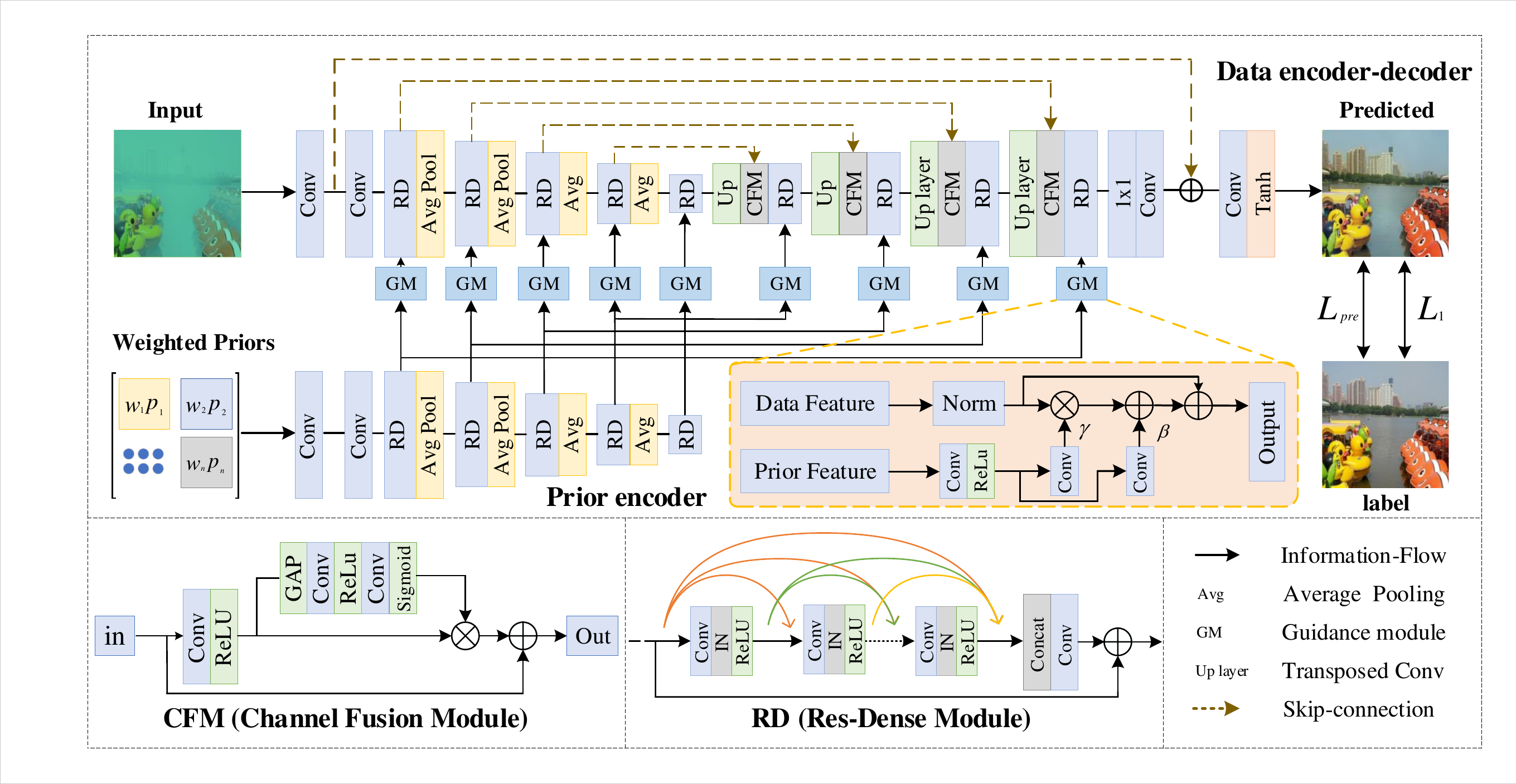}}
	\caption{Overview of our synthesis network structure. Synthesis network consists of a prior encoder and a data encoder-decoder. The prior encoder receives weighted priors to extract prior features and the data encoder-decoder takes the original image as input to extract data features. Then, prior features are introduced into the data encoder-decoder by a prior guidance module to assist image enhancement.}
	\label{Synthesis Network}
\end{figure*}

The detailed structure of our proposed adaptive weighting module is shown in Fig.\ref{Analysis Network}, which is achieved by a SENet architecture~\cite{Hu2018}. Denote the input image and each prior as $Input$ and $P_{i}, i\in\{0,1,2,...,n\}$, respectively. We first process them by a separate 1x1 convolution with 16 channels to ensure that the number of each input channel keeps the same, avoiding unbalanced learning of the adaptive weighting module, i.e, $d_{i}=16$. The module then applies the global average pooling layer (GAP) at input and each prior maps to squeeze their spatial information. Then, the feature maps of the input image and each prior are concatenated together, obtaining the aggregated information ${P}_{D}$, i.e, ${P}_{D}=\left\{Input; P_{1}; \ldots ; P_{n}\right\}$, where $D=\sum_{i=0}^{n} d_{i}$, $n$ is the number of priors. The module adopts two fully connected (FC) layers to process the features ${P}_{D}$ with a Rectified Linear Unit non-linear (ReLU) function after each layer. The last layer is a sigmoid function, which predicts a set of weight maps $\left\{W_{1}; W_{2}; \ldots ; W_{n}\right\}$. The weight of each prior is a one-dimensional parameter, $W_{n} \in[\mathrm{B}, 1,1,1]$, in which B denotes the batch-size, i.e.,
\begin{equation}
\label{eq15}
\left\{W_{1} ; W_{2} ; \ldots ; W_{n}\right\}=D_{\operatorname{conv}}\left(P_{D} ; \theta\right)
\end{equation}
where $D_{conv}$ is the  proposed adaptive weighting module. With the fusion weight $W_{n}$, a set of dynamically weighted priors $\left\{P_{1} W_{1} ; P_{2} W_{2} ; \ldots ; P_{n} W_{n}\right\}$ are presented as the input of synthesis network to guide image enhancement. As described above, three priors are chosen in this paper, e.g, $n = 3$.

\subsection{Synthesis Network}
Synthesis network aims to effectively aggregate priors and data information to perform underwater image enhancement, which consists of two sub-networks: a prior encoder and a data encoder-decoder. The entire architecture of synthesis network is shown in Fig.\ref{Synthesis Network}. Specifically, the prior encoder and the data encoder-decoder are designed to extract prior features and data features, respectively. In our implementation, the prior encoder is designed to be with the same structure of the data encoder. The shallow and deep features from the prior encoder are introduced into each layer of the data encoder and decoder respectively. Such an architecture design can implicitly benefit each other in a layer-by-layer manner. 

A novel prior guidance module (GM) achieved by SFT layer~\cite{wang2018recovering} is designed to effectively introduce priors into the data, which holds two advantages. One is to regularize data information to avoid overfitting of synthesized images, which is achieved by normalizing data feature maps. Another is to guide the data encoder-decoder to perform image enhancement and better optimize the network, which is accomplished by learning a pair of prior modulation parameters, $\gamma$ and $\beta$. 

Specifically, as shown in the GM of Fig.\ref{Synthesis Network}, data feature maps, $\mathrm{S} \in R^{H W C}$, are normalized using instance normalization~\cite{ulyanov2016instance}, where $H$, $W$ and $C$ denote the height of feature maps, the width of feature maps and the number of channels, respectively. The normalization way can be expressed as:
\begin{equation}
\label{eq21}
\hat{S}=\frac{S-\mu_{c}}{\sigma_{c}}
\end{equation}
where $\mu_{c}$ and $\sigma_{c}$ are the mean and standard deviation of $S$ in channel $C$. 

Prior features are also transformed into two learnable prior parameters to fine-tune data features, namely $(\gamma, \beta) = G(\Phi_{w})$, where $G$ represents the learning network. In our implementation, a 3 × 3 convolution with 64 channels is first adopted, followed by an activation function ReLU, to process the prior features from the prior encoder. Two 3 × 3 convolutions with 64 channels are used to generate $\gamma, \beta$ in a pixel-wisely manner, respectively. Finally, by fusing the priors and data, the output of prior guidance module is, 
\begin{equation}\label{eq22}
\hat{S}^{\text {New}}=\hat{S} * (\gamma + 1) + \beta
\end{equation}
where $\gamma$ and $\beta$ can be regarded as the self-adaptive scale factor and bias factor, respectively, holding the same shape of the data feature map $S$. By learning the pair of parameters through prior information, different channels and pixels will obtain different scales and bias factors, and more attention will be paid to useful information. In addition, $(\gamma + 1)$ also introduces a residual branch to better maintain data information and improve the gradient flow.

\subsection{Training Losses}
Taking both pixel-level and feature-level loss into consideration, the reconstruction loss and perceptual loss are adopted in our model. The reconstruction loss is defined as follows,
\begin{equation}\label{eq24}
L_{rec}=\left\|I^{\text {out}}-I^{gt}\right\|_{1}
\end{equation}
where $I^{\text {out}}$ and $I^{g t}$ denote the enhanced and ground-truth image, respectively. The pixel-level reconstruction loss is conducive to content fidelity. However, it is limited in capturing high-level semantics, which is not consistent with human perception. 

The perceptual loss is defined as the sum of distances between the designated output features of pre-trained VGG-19 network on ImageNet, which is written as:
\begin{equation}\label{eq25}
L_{per}=\frac{1}{H * W} \sum_{i=1}^{H} \sum_{j=1}^{W}\left[\varphi\left(I^{\text {out}}\right)_{i, j}-\varphi\left(I^{gt}\right)_{i, j}\right]^{2}
\end{equation}
where $\varphi(\cdot)$ is the feature map of the pool-3 layer of the pre-trained VGG-19 network. $H$, $W$ are the height and width of feature maps. The overall loss function is defined as follows, 
\begin{equation}\label{eq26}
L=\lambda_{1} L_{rec}+\lambda_{2} L_{per}
\end{equation}
where $\lambda_{1}$ and $\lambda_{2}$ are trade-off weights. In our work, they are set as $\lambda_{1}$ = 0.8 and $\lambda_{2}$ = 0.2.

\section{Experimental results}
In this section, the implementation details and experiment settings of our framework are first presented. Then, we evaluate the proposed underwater synthetic dataset and analyze the experimental results of on both synthetic and real underwater images. Finally, a series of ablation studies are provided to verify each component of our ANA-SYN and the model complexity and running time are analyzed.

\begin{figure*}[!t]
	\centering
	\centerline{\includegraphics[width=17.8cm,height=5.8cm]{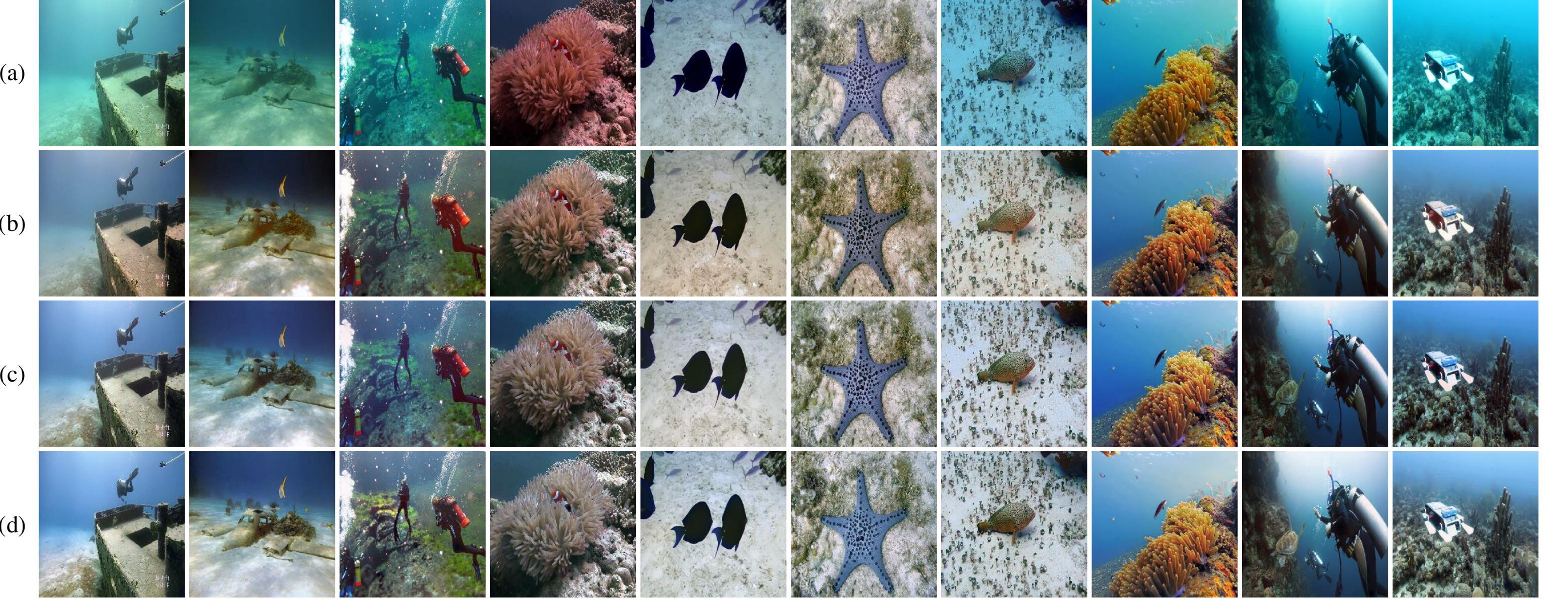}}
	\caption{Enhancement results of U-net trained on different synthetic datasets. (a): underwater images sampled from~\textbf{UFO-120}; (b): enhanced results using the model trained by the synthetic dataset based on~\cite{Li2020UnderwaterSP}; (c): enhanced results using the model trained by the synthetic dataset based on~\cite{ding2019jointly}; (d): enhanced results using the model trained by our proposed synthetic dataset. More results are shown in the supplementary material.}
	\label{Data_test}
\end{figure*}

\begin{table*}[!t]
	\setlength{\abovecaptionskip}{0cm}
	\setlength{\belowcaptionskip}{-0.2cm}
	\centering
	\footnotesize
	\caption{Quantitative results (average uiqm/niqe) of different synthetic datasets on real benchmarks (RUIE, SQUID, UIEB, EUVP, and UFO-120). In this paper, we highlight the top one performance in red.}
	\renewcommand{\arraystretch}{1.2}
	\begin{tabular}{c|c|c|c|c|c|c|c|c|c|c}
		\hline
		\multirow{2}{*}{Methods}&\multicolumn{5}{c|}{UIQM$\uparrow$} &\multicolumn{5}{c}{NIQE$\downarrow$} \\\cline{2-11}
		&RUIE &SQUID &UIEB &EUVP &UFO-120  &RUIE&SQUID &UIEB &EUVP &UFO-120 \\\hline
		\hline
		\makecell[c]{U-net based on~\cite{Li2020UnderwaterSP}}  &3.080&2.369 &3.091&3.234 &3.061 &4.897 &8.661 &4.869&7.886 &6.280 \\
		\makecell[c]{U-net based on~\cite{ding2019jointly}}  &{\color{red}{3.137}}&{\color{red}{2.547}}&3.122&3.249 &3.081&{\color{red}{4.506}}&{\color{red}{7.298}}&4.646&7.726&5.963 \\
		\makecell[c]{U-net based on our dataset} &3.109&2.513 &{\color{red}{3.135}}&{\color{red}{3.260}} &{\color{red}{3.092}}&4.522&7.952&{\color{red}{4.577}}&{\color{red}{7.545}}&{\color{red}{5.878}}\\\hline
		\makecell[c]{UGAN based on~\cite{Li2020UnderwaterSP}}  &3.034&2.960&3.122&3.136&3.111&6.905 &6.759 &6.418&6.853 &6.316 \\
		\makecell[c]{UGAN based on~\cite{ding2019jointly}}  &3.141&3.016&3.247&3.278&3.223&5.015 &5.486 &4.935&{\color{red}{5.957}}&5.187 \\
		\makecell[c]{UGAN based on our dataset}  &{\color{red}{3.177}}&{\color{red}{3.064}}&{\color{red}{3.267}}&{\color{red}{3.314}}&{\color{red}{3.271}}&{\color{red}{4.910}}&{\color{red}{5.479}}&{\color{red}{4.886}}&6.097&{\color{red}{5.181}}\\\hline
		\makecell[c]{ANA-SYN based on~\cite{Li2020UnderwaterSP}}  &3.088&2.364&3.121&3.247 &3.094&4.925&9.095&4.900&7.650&6.079 \\
		\makecell[c]{ANA-SYN based on~\cite{ding2019jointly}} &3.128&2.478&3.140&{\color{red}{3.259}}&3.096&4.763&8.360&4.794&7.892&6.061 \\
		\makecell[c]{ANA-SYN based on our dataset}  &{\color{red}{3.158}}&{\color{red}{2.617}}&{\color{red}{3.151}}&3.250&{\color{red}{3.105}}&{\color{red}{4.542}}&{\color{red}{6.216}}&{\color{red}{4.608}}&{\color{red}{7.255}}&{\color{red}{5.737}}\\\hline\hline
	\end{tabular}
	\label{table1}
\end{table*}

\subsection{Implementation Details}
For training, we select synthetic images of nine types\footnote[2]{{Type I, IA, IB, II and III for open ocean water and type 1C, 3C, 5C and 7C for coastal water}} of water except type 9C since it is too turbid. The dataset (9000 pairs) is divided into two groups, containing 6300 (700x9) pairs of training images, denoted as~\textbf{Train-S6300}, and 2700 pairs of testing (300x9) images, denoted as~\textbf{Test-S2700}. All input images are resized to 256 × 256 and their pixel values are normalized to $[-1,1]$. In addition, the training examples are augmented by randomly rotating $90^\circ, 180^\circ, 270^\circ$ and horizontal flipping. 

Our proposed ANA-SYN is implemented using the Pytorch tool and trained for 100 epochs on two NVIDIA Titan V GPUs. Adam is used as the optimization algorithm with a mini-batch size of 6 and the learning rate is fixed as $2e^{-4}$. Default values of $\beta_{1}$ and $\beta_{2}$ are set as 0.5 and 0.999, respectively. The weight decay is set to 0.00005.

\subsection{Experiment Settings}
For testing, we compare our method with seven state-of-the-art underwater image enhancement methods on both synthetic data~\textbf{Test-S2700} and five real underwater benchmarks,~\textbf{RUIE}\footnote[3]{{The RUIE dataset contains 3930 real-world underwater images}}~\cite{liu2019real},~\textbf{SQUID}\footnote[4]{The SQUID dataset contains 57 real-world underwater images}~\cite{Berman2020UnderwaterSI},~\textbf{UIEB}\footnote[5]{The UIEB dataset contains 950 real-world underwater images}~\cite{li2019underwater},~\textbf{EUVP}\footnote[6]{The EUVP dataset contains 1910 real-world underwater images}~\cite{islam2020fast} and~\textbf{UFO-120}\footnote[7]{The UFO-120 dataset contains 3255 real-world underwater images}~\cite{2019IslamToward}. The compared methods include traditional methods (Fusion~\cite{ancuti2012enhancing}, UIBLA~\cite{peng2017underwater} and HE-Prior~\cite{li2016underwater}) and deep-learning methods (UGAN~\cite{fabbri2018enhancing}, FUIE-GAN~\cite{islam2020fast}, Water-Net~\cite{li2019underwater} and Ucolor~\cite{Li2021}).

For synthetic data~\textbf{Test-S2700}, we retrain all deep-learning methods on~\textbf{Train-S6300} and evaluate them using the full-reference metrics Peak Signal to Noise Ratio (PSNR) and Structural Similarity (SSIM). For PSNR, the higher scores mean the result is closer to the image content of the ground truth. For SSIM, the larger values denote that the result is more similar to the structure and texture of the ground truth. 

For five real underwater benchmarks, all competing methods are tested using the corresponding test models and parameters released by their authors. We employ the no-reference metrics UIQM and NIQE scores as references to compare the performance of different methods since these real underwater images do not have corresponding ground-truths. A higher UIQM or a lower NIQE score represents a better human visual perception. It should be pointed out that these scores do not accurately reflect the performance of various underwater enhancement methods in some cases~\cite{li2019underwater, Li2021}.

We also take a user study to more accurately score the performance of different methods on five real benchmarks, denoted as “Perceptual Scores”. Specifically, 12 participants having experience in image processing are invited to score enhanced results. Each participant has no time limitation and does not know the result produced by which method. The scores have 5 levels ranging from 1 to 5, indicating worst to best quality. In addition, we measure the color restoration accuracy on the 16 representative examples presented in the project page\footnote[8]{http://csms.haifa.ac.il/profiles/tTreibitz/datasets/ambient\_forwardlooking/\\index.html} of SQUID~\cite{Berman2020UnderwaterSI} to evaluate the color accuracy of different methods.

\subsection{Dataset Evaluation}
In this part, we aim to evaluate the effectiveness of our proposed synthetic dataset by retraining the compared methods. Three networks (U-net, UGAN and ANA-SYN) are selected and trained on our proposed dataset and other synthetic datasets~\cite{Li2020UnderwaterSP,ding2019jointly}, respectively. The number and content of training samples remain the same and only the synthesis model is different. The trained networks are tested on five real underwater benchmarks. 

Due to the limited space, only the visual comparisons results of U-net on UFO-120 dataset are presented in Fig.\ref{Data_test} and more experimental results are given in the supplementary material. It can be observed that the network trained on our proposed dataset not only has a good performance in color correction and contrast enhancement but also enhances details, achieving a higher generalization ability in real underwater data. 

Additionally, we perform quantitative comparisons on five real underwater benchmarks using UIQM and NIQE metrics and the results are presented in Table~\ref{table1}. As presented, the three models trained on our dataset almost achieve the best results in term of UIQM and NIQE metrics. Such results also demonstrate the superiority of our proposed dataset.
\label{section:my}

\subsection{Network Architecture Evaluation}
In this section, we perform quantitative and visual comparisons on both synthetic and real underwater images.

\begin{figure*}[!t]
	\centering
	\centerline{\includegraphics[width=18.1cm, height=3.4cm]{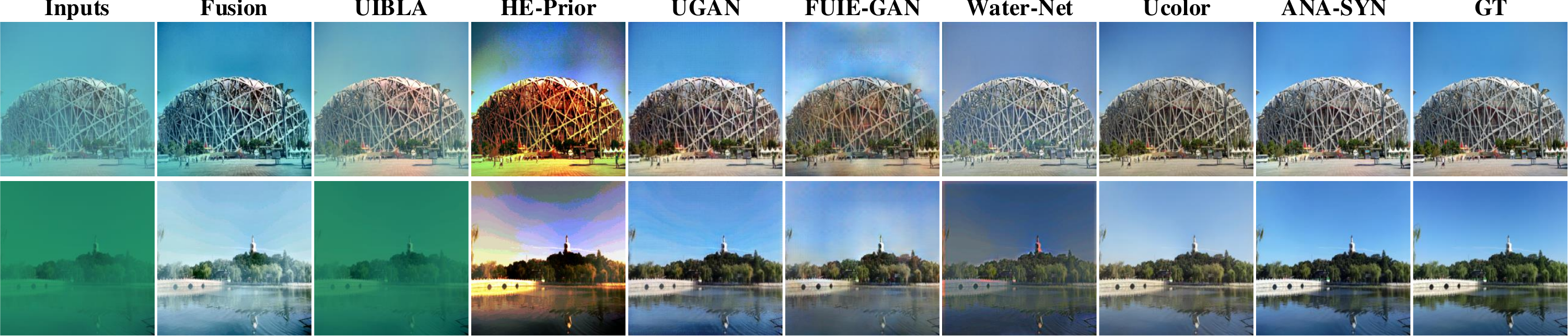}} 
	\caption{Visual comparisons on challenging underwater images sampled from~\textbf{Test-S2700}. From left to right are raw images, and the results of Fusion~\cite{ancuti2012enhancing}, UIBLA~\cite{peng2017underwater}, HE-Prior~\cite{li2016underwater}, UGAN~\cite{fabbri2018enhancing}, FUIE-GAN~\cite{islam2020fast}, Water-Net~\cite{li2019underwater}, Ucolor~\cite{Li2021}, our ANA-SYN and ground truths.}
	\label{Test-S2700} 
\end{figure*}

\begin{table*}[!t]
	\setlength{\abovecaptionskip}{0cm}
	\setlength{\belowcaptionskip}{-0.2cm}
	\centering
	\footnotesize
	\caption{Quantitative results (average psnr (db)/ssim) of different methods on synthetic testing set (Test-S2700). In this paper, we highlight the top one performance in red.}
	\renewcommand{\arraystretch}{1.2}
	\begin{tabular}{c|c|c|c|c|c|c|c|c}
		\hline\hline
		Methods &Fusion~\cite{ancuti2012enhancing}&UIBLA~\cite{peng2017underwater} & HE-Prior~\cite{li2016underwater} &UGAN~\cite{fabbri2018enhancing} &FUIE-GAN~\cite{islam2020fast}&Water-Net~\cite{li2019underwater}  & Ucolor~\cite{Li2021} &Our ANA-SYN \\\hline
		\makecell[c]{PSNR} &13.496 &12.059 &14.767&23.393&19.373&17.114&24.189&{\color{red}{26.869}}\\
		\makecell[c]{SSIM} &0.774 &0.642 &0.723&0.903 &0.836&0.821&0.947&{\color{red}{0.963}} \\
		\hline\hline
	\end{tabular}
	\label{table2}
\end{table*}

\begin{table*}[!t]
	\setlength{\abovecaptionskip}{0cm}
	\setlength{\belowcaptionskip}{-0.2cm}
	\centering
	\footnotesize
	\caption{Quantitative results (average uiqm/niqe) of different methods on real benchmarks (RUIE, SQUID, UIEB, EUVP and UFO-120). The top three results are marked in red, blue and green. '-' represents the results are not available.}
	\renewcommand{\arraystretch}{1.2}
	\begin{tabular}{c|c|c|c|c|c|c|c|c|c|c|c|c}
		\hline
		\multirow{2}{*}{Methods}&\multicolumn{6}{c|}{UIQM$\uparrow$} &\multicolumn{6}{c}{NIQE$\downarrow$} \\\cline{2-13}
		&RUIE&SQUID &UIEB &EUVP &UFO-120 &Avg &RUIE&SQUID &UIEB &EUVP &UFO-120 &Avg\\\hline
		\hline
		\makecell[c]{Fusion~\cite{ancuti2012enhancing}} &3.048&1.991&2.870 &2.779&2.769&2.691&{\color{red}{4.250}}&{\color{red}{4.767}}&{\color{red}{4.300}}&8.853 &{\color[rgb]{0,0.5,0}{5.606}}&{\color{blue}{5.555}}\\
		\makecell[c]{UIBLA~\cite{peng2017underwater}} &2.378&1.047&2.262&2.053&2.061&1.960&5.024&7.350&4.689&8.157 &6.099&6.264\\
		\makecell[c]{HE-Prior~\cite{li2016underwater}} &---&--- &2.637 &2.656&2.565&2.619&{\color{blue}{4.269}}&{\color{blue}{5.902}}&{\color{blue}{4.371}}&9.165 &5.986&5.939\\
		\makecell[c]{UGAN~\cite{fabbri2018enhancing}} &{\color{red}{3.192}}&{\color{red}{2.927}}&{\color{red}{3.180}}&{\color{blue}{3.237}}&{\color{red}{3.152}}&{\color{red}{3.138}}&5.474&6.253&4.705&{\color{red}{5.730}}&{\color{red}{4.978}}&{\color{red}{5.428}}\\
		\makecell[c]{FUIE-GAN~\cite{islam2020fast}} &3.002&1.775&2.949&3.079&2.954&2.752&5.913&7.736&5.745&7.818&6.000&6.642 \\
		\makecell[c]{Water-Net~\cite{li2019underwater}} &{\color[rgb]{0,0.5,0}{3.081}}&{\color[rgb]{0,0.5,0}{2.316}}&3.002&3.052&2.908&2.872&5.164&7.244&5.313&{\color[rgb]{0,0.5,0}{6.992}}&5.821&6.107 \\
		\makecell[c]{Ucolor~\cite{Li2021}} &2.984&2.292&{\color[rgb]{0,0.5,0}{3.048}}&{\color[rgb]{0,0.5,0}{3.171}}&{\color[rgb]{0,0.5,0}{3.001}}&{\color[rgb]{0,0.5,0}{2.899}}&5.304&7.314&5.004&{\color{blue}{6.395}}&{\color{blue}{5.523}}&5.908 \\
		Our ANA-SYN &{\color{blue}{3.158}}&{\color{blue}{2.617}}&{\color{blue}{3.151}}&{\color{red}{3.250}}&{\color{blue}{3.105}}&{\color{blue}{3.056}}&{\color[rgb]{0,0.5,0}{4.542}}&{\color[rgb]{0,0.5,0}{6.216}}&{\color[rgb]{0,0.5,0}{4.608}}&7.255&5.737&{\color[rgb]{0,0.5,0}{5.672}}\\
		\hline\hline
	\end{tabular}
	\label{table3}
\end{table*}

\begin{table*}[!t]
	\setlength{\abovecaptionskip}{0cm}
	\setlength{\belowcaptionskip}{-0.2cm}
	\centering
	\footnotesize
	\caption{Quantitative results (average color error/perceptual scores) of different methods on real benchmarks (RUIE, SQUID, UIEB, EUVP and UFO-120). The top three results are marked in red, blue and green.}
	\renewcommand{\arraystretch}{1.2}
	\begin{tabular}{c|c|c|c|c|c|c|c|c|c|c|c}
		\hline
		\multirow{2}{*}{Methods}&\multicolumn{5}{c|}{Color Error$\downarrow$} &\multicolumn{6}{c}{Perceptual Scores$\uparrow$} \\\cline{2-12}
		&Katzaa&Michmoret &Nachsholim &Satil &Avg  &RUIE& SQUID &UIEB &EUVP &UFO-120 &Avg \\\hline
		\hline
		\makecell[c]{Fusion~\cite{ancuti2012enhancing}} &31.161&28.246 &31.985 &35.220 &31.653 &2.483&2.272 &2.624 &2.483 &2.616 &2.496\\
		\makecell[c]{UIBLA~\cite{peng2017underwater}}&34.214&32.742 &32.559 &36.019 &33.883 &2.104&1.959 &2.059 &2.180 &2.105 &2.081\\
		\makecell[c]{HE-Prior~\cite{li2016underwater}} &21.586&{\color[rgb]{0,0.5,0}{17.526}} &{\color{blue}{10.542}}&26.634 &19.072 &1.782&1.091 &1.724 &2.083 &1.902 &1.716\\
		\makecell[c]{UGAN~\cite{fabbri2018enhancing}} &{\color{red}{10.158}}&{\color{red}{11.714}} &{\color{red}{10.206}} &{\color{blue}{7.650}} &{\color{red}{9.932}}&1.389&1.575 &1.519 &1.803 &1.561 &1.569\\
		\makecell[c]{FUIE-GAN~\cite{islam2020fast}} &26.847&25.735&24.820 &29.663 &26.766 &2.672&2.313 &2.518 &2.542 &2.744 &2.558\\
		\makecell[c]{Water-Net~\cite{li2019underwater}} &23.352&21.695 &20.438 &22.190 &21.918 &{\color{blue}{2.987}}&{\color[rgb]{0,0.5,0}{2.482}}&{\color[rgb]{0,0.5,0}{2.764}}&{\color[rgb]{0,0.5,0}{2.722}}&{\color[rgb]{0,0.5,0}{2.847}}&{\color[rgb]{0,0.5,0}{2.760}}\\
		\makecell[c]{Ucolor~\cite{Li2021}} &{\color[rgb]{0,0.5,0}{15.782}}&18.189 &22.456 &{\color[rgb]{0,0.5,0}{13.820}} &{\color[rgb]{0,0.5,0}{17.562}}&{\color[rgb]{0,0.5,0}{2.979}}&{\color{blue}{2.626}}&{\color{blue}{2.909}}&{\color{blue}{3.011}} &{\color{red}{2.913}} &{\color{blue}{2.888}}\\
		Our ANA-SYN &{\color{blue}{10.413}} &{\color{blue}{15.550}} &{\color[rgb]{0,0.5,0}{11.297}} &{\color{red}{5.847}} &{\color{blue}{10.776}} &{\color{red}{3.038}} &{\color{red}{3.763}} &{\color{red}{3.042}}&{\color{red}{3.013}} &{\color{blue}{2.902}}&{\color{red}{3.152}}\\
		\hline\hline
	\end{tabular}
	\label{table4}
\end{table*}

\subsubsection{\textbf{Experiment on Synthetic Datasets}}
Some quantitative comparisons are conducted on~\textbf{Test-S2700} and the corresponding results are reported in Table~\ref{table2}. It can be observed that our ANA-SYN achieves the best performance in term of PSNR and SSIM metrics and clearly surpasses other state-of-the-art methods by a wide margin. Such results demonstrate that the effective integration of priors and data information can better enhance underwater images. For Water-Net, its performance is limited by three pre-processing versions, where the white balance method they used cannot pre-process the input data well, resulting in a low PSNR value. For FUIE-GAN, its purpose is to achieve a fast and lightweight model with fewer parameters, which can easily reach performance bottlenecks on the complex and distorted training samples. UGAN is trained in an end-to-end manner without introducing underwater priors and its performance in two metrics is relatively poor. Ucolor only introduces the transmission map prior and does not adaptively weight multiple priors based on their usefulness for images, leading to a limited performance.

Some visual comparison results on synthetic images sampled from~\textbf{Test-S2700} are also shown in Fig.\ref{Test-S2700}. It can be observed that the enhanced results generated by our method have clear details with fewer color artifacts and high-fidelity object regions, which are closer to the ground-truths. In comparison, the competing methods tend to introduce color artifacts or fail to remove haze on images. He-Prior exhibits some serious color artifacts, producing visually over-enhancement results. Fusion and UIBLA can improve the contrast of images to some extent, but they cannot effectively remove color shift and fail to recover the complete scene structure. Although Water-Net, FUIE-GAN, UGAN and Ucolor can provide a relatively good color appearance, they often suffer from local over-enhancement and remain some haze in the results.

\begin{figure*}[!t]
	\centering
	\centerline{\includegraphics[scale = 0.54]{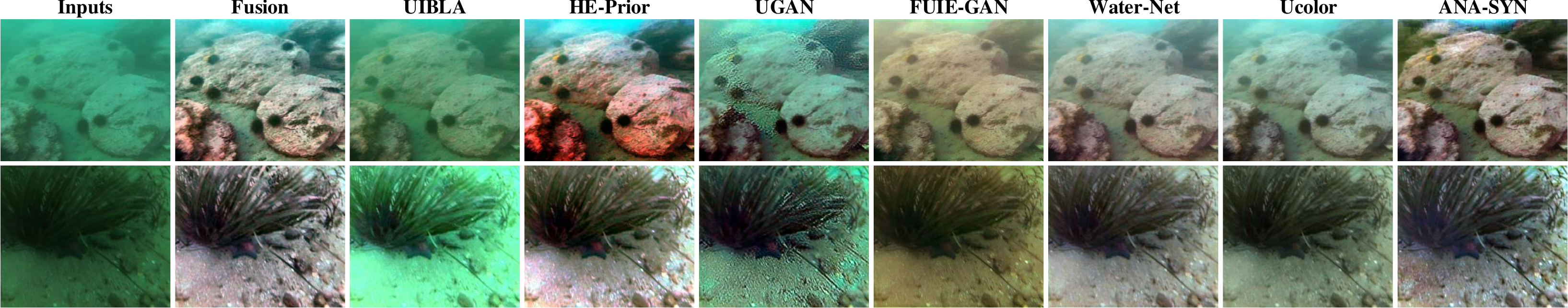}} 
	\caption{Visual comparisons on challenging underwater images sampled from~\textbf{RUIE}. From left to right are raw images, and the results of Fusion~\cite{ancuti2012enhancing}, UIBLA~\cite{peng2017underwater}, HE-Prior~\cite{li2016underwater}, UGAN~\cite{fabbri2018enhancing}, FUIE-GAN~\cite{islam2020fast}, Water-Net~\cite{li2019underwater}, Ucolor~\cite{Li2021} and our ANA-SYN. More results are given in supplementary material.}
	\label{RUIE} 
\end{figure*}

\begin{figure*}[!t]
	\centering
	\centerline{\includegraphics[scale = 0.54]{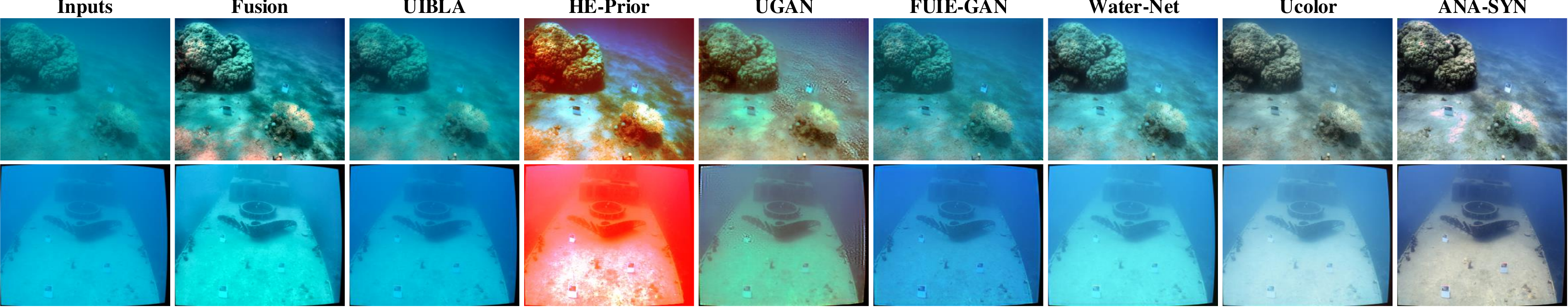}} 
	\caption{Visual comparisons on challenging underwater images sampled from~\textbf{SQUID}. From left to right are raw images, and the results of Fusion~\cite{ancuti2012enhancing}, UIBLA~\cite{peng2017underwater}, HE-Prior~\cite{li2016underwater}, UGAN~\cite{fabbri2018enhancing}, FUIE-GAN~\cite{islam2020fast}, Water-Net~\cite{li2019underwater}, Ucolor~\cite{Li2021} and our ANA-SYN. More results are shown in supplementary material.}
	\label{SQUID} 
\end{figure*}

\begin{figure*}[!t]
	\centering
	\centerline{\includegraphics[scale = 0.54]{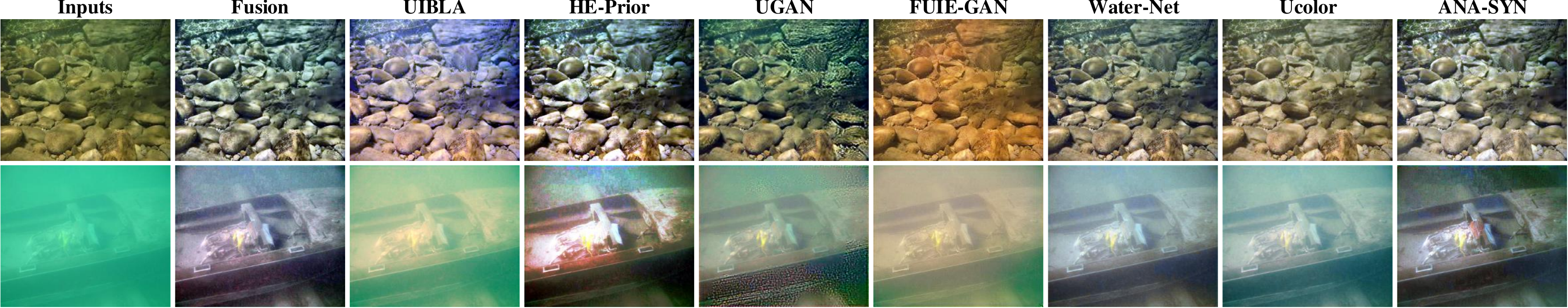}} 
	\caption{Visual comparisons on challenging underwater images sampled from~\textbf{UIEB}. From left to right are raw images, and the results of Fusion~\cite{ancuti2012enhancing}, UIBLA~\cite{peng2017underwater}, HE-Prior~\cite{li2016underwater}, UGAN~\cite{fabbri2018enhancing}, FUIE-GAN~\cite{islam2020fast}, Water-Net~\cite{li2019underwater}, Ucolor~\cite{Li2021} and our ANA-SYN. More results are shown in supplementary material.}
	\label{UIEB} 
\end{figure*}

\begin{figure*}[!t]
	\centering
	\centerline{\includegraphics[scale = 0.54]{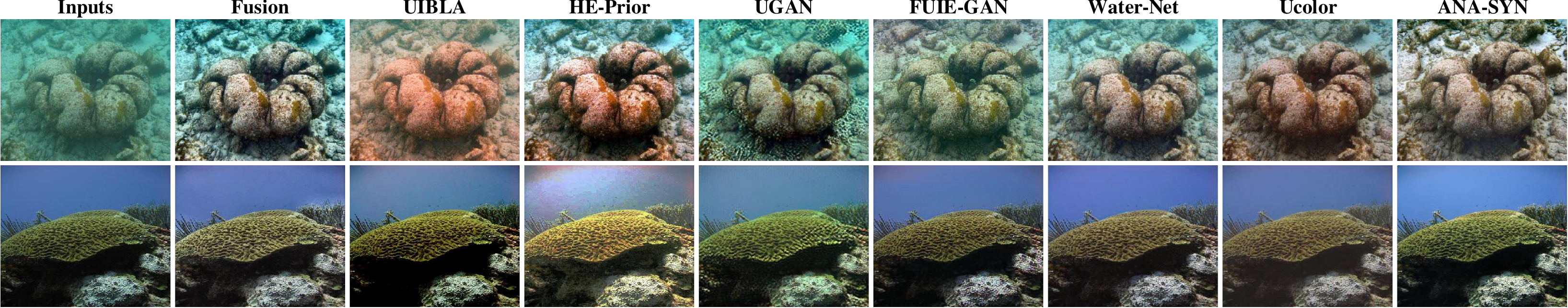}} 
	\caption{Visual comparisons on challenging underwater images sampled from~\textbf{EUVP}. From left to right are raw images, and the results of Fusion~\cite{ancuti2012enhancing}, UIBLA~\cite{peng2017underwater}, HE-Prior~\cite{li2016underwater}, UGAN~\cite{fabbri2018enhancing}, FUIE-GAN~\cite{islam2020fast}, Water-Net~\cite{li2019underwater}, Ucolor~\cite{Li2021} and our ANA-SYN. More results are shown in supplementary material.}
	\label{EUVP} 
\end{figure*}

\begin{figure*}[!t]
	\centering
	\centerline{\includegraphics[scale = 0.54]{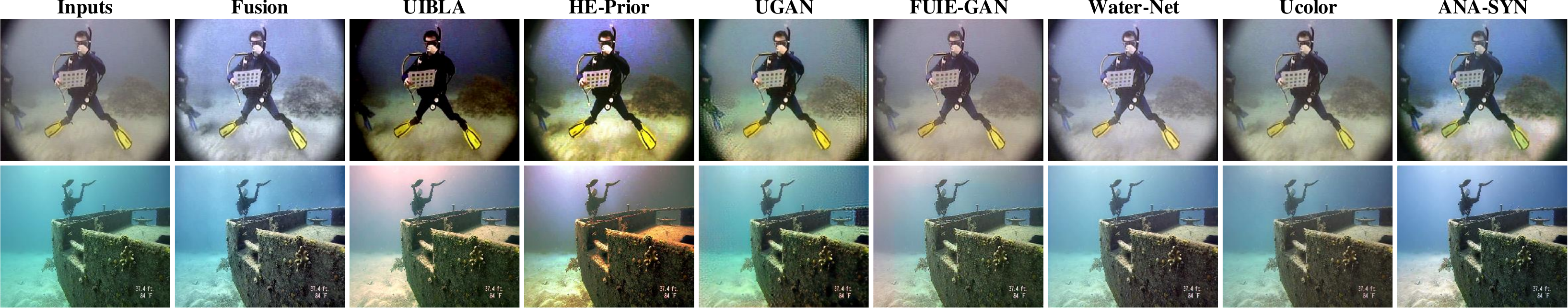}} 
	\caption{Visual comparisons on challenging underwater images sampled from~\textbf{UF0-120}. From left to right are raw images, and the results of Fusion~\cite{ancuti2012enhancing}, UIBLA~\cite{peng2017underwater}, HE-Prior~\cite{li2016underwater}, UGAN~\cite{fabbri2018enhancing}, FUIE-GAN~\cite{islam2020fast}, Water-Net~\cite{li2019underwater}, Ucolor~\cite{Li2021} and our ANA-SYN. More results are shown in supplementary material.}
	\label{UFO-120} 
\end{figure*}

\subsubsection{\textbf{Experiment on Real Datasets}}
To evaluate the generalization ability of our proposed method on real images, we conduct comprehensive experiments on five more underwater benchmarks. The average scores of UIQM and NIQE metrics are reported in Table~\ref{table3}, where it can be observed that our method achieves relatively higher UIQM and NIQE scores. For the UIQM metric, UGAN almost achieves the highest score on five real datasets, while our ANA-SYN ranks the second best. For the NIQE metrics, our method ranks the third best, only slightly lower than Fusion and UGAN. 

To further analyze the robustness of color accuracy, we report the average color error on SQUID in the left part of Table~\ref{table4}. For images in Satil dive sites, our method achieves the lowest color error. For images in Katzaa and Nachsholim dive sites, our results only have a little lower than UGAN. Observing the perceptual scores in the right part of Table~\ref{table4}, we can see that the learning-based methods achieve relatively higher performance. Among them, our ANA-SYN is significantly outperform other competing methods, obtaining the best performance. Such results demonstrate that our method can generalize well to different real-world underwater scenarios.

It is interesting that the UGAN method has similar or only slightly higher UIQM/NIQE/Color Error values in comparison to our ANA-SYN, but the perceptual score is far worse than ours. In our opinion, this is mainly caused by the fact that these metrics are biased to some characteristics (not entire image) and do not consider the color shift and artifacts, and thus they are not sensitive to artifacts generated on the boundary of objects~\cite{li2019underwater}. As shown in the enhanced results of UGAN in Fig.\ref{RUIE} to Fig.\ref{UFO-120}, we can clearly observe the edges of objects are blurred since the model employs data generated based on Cycle-GAN migration for training. 

We also present some results of different methods on images sampled from~\textbf{RUIE} and~\textbf{SQUID} in Fig.\ref{RUIE} and Fig.\ref{SQUID}. As shown, our method not only effectively removes the seriously greenish and blueish color tone but also restores clear structures, which is credited to the effective combination of priors and data. By contrast, UIBLA, UGAN, FUIE-GAN and Ucolor remain some unexpected color casts, UGAN even produces artifacts at the boundary of objects and He-Prior introduces over-enhancement in their results. 

Some enhanced results of different methods on~\textbf{UIEB} are also presented in Fig.\ref{UIEB}. For the image with yellowish tone in shallow water areas or the image with greenish tone in deep water areas, our method significantly corrects the color casts and dehazes the input image, which shows the advantage of our specially designed adaptive weighting module. In comparison, most comparison methods still suffer from obvious color artifacts and blurry details. 

Visual comparisons on challenging underwater images sampled from~\textbf{EUVP} and~\textbf{UFO-120} are shown in Fig.\ref{EUVP} and Fig.\ref{UFO-120}. For these low or high quality underwater images, our method can effectively enhance foregrounds structure, remit color casts and remove the haze on images, leading to more natural results and producing visually plausible results. By contrast, the competing methods often introduce annoying artifacts to outputs, destroy the image structure and introduce unexpected color casts in recovery results due to ignoring lots of favorable underwater priors. Due to the limit space, more comparative results in five real benchmarks are provided in supplementary material.

\begin{table}[!t]
	\setlength{\abovecaptionskip}{0cm}
	\setlength{\belowcaptionskip}{-0.2cm}
	\centering
	\footnotesize
	\caption{Quantitative results (average PSNR/SSIM/UIQM/NIQE) of the ablation study on synthetic and five real benchmarks.}
	\renewcommand{\arraystretch}{1.2}
	\begin{tabular}{c|c|c|c|c|c}
		\hline
		\multirow{2}{*}{Modules} &\multirow{2}{*}{Models} &\multicolumn{2}{c|}{Synthetic data} &\multicolumn{2}{c}{Real benchmarks} \\\cline{3-6}
		&&PSNR$\uparrow$ &SSIM$\uparrow$ &UIQM$\uparrow$&NIQE$\downarrow$  \\\hline
		\hline
		\makecell[c] &{~\textbf{full model}} &{\color{red}{26.869}} &{\color{red}{0.963}} &{\color{blue}{3.056}} &{\color{red}{5.672}}\\\hline
		\makecell[c] {\multirow{4}{*}{UPriors}}&{BL} &25.803 &0.955 &3.022 &6.095\\
		\makecell[c] &{BL+E} &26.415 &0.958 &{\color{blue}{3.056}} &5.824\\
		\makecell[c] &{BL+C} &26.536 &0.961 &3.053 &{\color{blue}{5.742}}\\
		\makecell[c] &{BL+W} &26.758 &{\color{blue}{0.962}} &3.035 &5.827\\\hline
		\makecell[c] {\multirow{2}{*}{AWM}}&{BL+A-CWE-G} &26.419 &0.959 &3.050 &5.857\\
		\makecell[c] &{BL+C-CWE-G} &26.464 &0.958 &3.051 &5.918\\\hline
		\makecell[c] {\multirow{2}{*}{GM}}&{BL+W-CWE-A} &26.717 &0.960 &{\color{red}{3.058}} &5.783\\
		\makecell[c] &{BL+W-CWE-C} &{\color{blue}{26.818}} &0.960 &3.046 &5.847\\
		\hline\hline
	\end{tabular}
	\label{table5}
\end{table}

\begin{table}[!t]
	\setlength{\abovecaptionskip}{0cm}
	\setlength{\belowcaptionskip}{-0.2cm}
	\centering
	\footnotesize
	\caption{Quantitative results (average psnr/ssim/perceptual score) of loss function on synthetic and five real benchmarks.}
	\renewcommand{\arraystretch}{1.2}
	\begin{tabular}{c|c|c|c}
		\hline
		Methods &\multicolumn{2}{c|}{Synthetic data} &\multicolumn{1}{c}{Real benchmarks}\\\hline
		&PSNR$\uparrow$ &SSIM$\uparrow$ &Perceptual Score$\uparrow$\\\hline
		\makecell[c]{w/o perc loss} &24.995 &0.931 &3.085 \\
		\makecell[c]{w/ perc loss} &26.869 &0.963 &3.152 \\
		\hline
	\end{tabular}
	\label{table6}
\end{table}

\begin{table}[!t]
	\setlength{\abovecaptionskip}{0cm}
	\setlength{\belowcaptionskip}{-0.2cm}
	\centering
	\footnotesize
	\caption{The flops, parameters and running time (image size is 256*256) of different deep methods on a pc with an intel(r) i5-10500 cpu, 16.0gb ram, a nvidia geforce rtx 2080 Super.}
	\renewcommand{\arraystretch}{1.2}
	\setlength{\tabcolsep}{1.7mm}{
		\begin{tabular}{c|c|c|c|c|c}
			\hline
			& Our &Ucolor &UGAN  &FUIE-GAN &Water-Net  \\ \hline
			Flops (G) &268.7 &43365 &3887 & 0.008 & 1937  \\ \hline
			Parameters (M) &110.1 &157.4 &57.17 & 4.216 & 1.091  \\ \hline		
			Running time (S) &0.097 &1.345 &0.009 & 0.081 & 0.582  \\ 
			\hline
		\end{tabular}
		\label{table7}}
\end{table}

\subsection{Ablation Study and Analysis}
To prove the effectiveness of each component, we conduct a series of ablation studies on Test-S2700 and five real benchmarks using PSNR/SSIM and UIQM/NIQE metrics. Four factors are mainly considered including the priors (UPriors), the adaptive weighting module (AWM), the prior guidance module (GM) and loss function, as follows:
\begin{itemize}
\item full model: recalibrating three priors by our proposed adaptive weighting module and using the prior guidance module to fuse. 
\item BL: a baseline U-net network without priors.
\item BL+E: adding the structure prior by the proposed prior guidance module.
\item BL+C: adding the color tone prior by the proposed prior guidance module.
\item BL+W: adding the water-type prior by the proposed prior guidance module. 
\item BL+A-CWE-G: directly adding three priors as the final priors and using the prior guidance module to fuse. 
\item BL+C-CWE-G: directly concatenating three priors as the final priors and using the prior guidance module to fuse.
\item BL+W-CWE-A: calibrating three priors by our proposed adaptive weighting module and using summation to fuse. 
\item BL+W-CWE-C: calibrating three priors by our proposed adaptive weighting module and using concatenation to fuse. 
\end{itemize}

The quantitative results are reported in Table~\ref{table5}. The models BL, BL+E, BL+C and BL+W can be employed to prove the effectiveness of priors, where it can be observed that the color tone, water-type and structure priors improve the overall performance by approximately 0.733dB, 0.955dB and 0.615dB, respectively. The average UIQM and NIQE scores of five real benchmarks are also relatively good overall. Such results also show the usefulness of priors for enhancement. 

The models BL+A-CWE-G and BL+C-CWE-G can be used to analyze the effectiveness of the proposed adaptive weighting module. Obviously, the results show that a straightforward method (concatenation or summation) cannot adaptively combine multiple priors into more discriminative priors according to the needs of the input image, which also shows the importance of our proposed module. 

As presented in BL+W-CWE-A and BL+W-CWE-C, compared with the direct concatenation or summation of prior features and data features, our proposed prior guidance module can improve performance by 0.05-0.15dB and generate more natural results (see UIQM and NIQE metrics). Such results demonstrate the superiority of our guidance module.

The performances of ANA-SYN with/without the perceptual loss are also compared in Table~\ref{table5}, where it is observed that the network that adds the perceptual loss in the training phase can restore more realistic colors (see PSNR and SSIM metrics on synthetic data) and improve the visual quality of final results (see the average perceptual scores on five real benchmarks), which is better than that trained without the perceptual loss.

\subsection{Model Complexity Analysis}
Table~\ref{table7} reports the parameters, flops and running time of our ANA-SYN and other representative deep methods. All methods are tested on a PC with an Intel(R) i5-10500 CPU, 16.0GB RAM, a NVIDIA GeForce RTX 2080 Super and the test image size is 256×256×3. Among them, our method achieves relatively few parameters, flops and running time. Although the flops, sizes and time cost of FUIE-GAN are smaller than ours, the generation capability in real underwater scenarios is far less than our method, even it is trained on real data (EUVP and UFO-120), see ‘Experiment on Real Datasets’ above. UGAN obtains the least running time across different methods. However, it has more flops and the performance in real images is limited. The parameters of Water-Net are 1.091M less than us, but its flops 1937G, which are larger than ours. For Ucolor, its parameters, flops and time cost far exceed ours. Such results demonstrate that our ANA-SYN can achieve superior performance with moderate parameters.

\section{Conclusion}
In this paper, a new underwater synthetic dataset is first proposed, in which a revised ambient light synthesis equation that defines the relationship among the ambient light values of RGB channels and many dependencies is embedded. Extensive evaluation results show that our proposed dataset is closer to real underwater data, which can be used as training data for various enhancement and restoration algorithms on underwater vision applications. Secondly, a new framework, called ANA-SYN, is proposed for underwater image enhancement under collaborations of priors (underwater domain knowledge) and data information (underwater distortion distribution). In addition, a novel adaptive weighting module is designed to adaptively calibrate priors according to the importance of each prior for the input image, and a new prior guidance module is introduced to effectively fuse prior and data features. Extensive experimental results on both synthetic data and five publicly available real underwater benchmarks demonstrate the effectiveness of our ANA-SYN.

\bibliographystyle{ieeetr}
\bibliography{ANA_SYN}
\end{document}